\documentclass{article}

\usepackage[preprint,nonatbib]{neurips_2020}

\usepackage{tikz}

\usepackage{tcolorbox}

\usepackage{graphicx}
\usepackage[utf8]{inputenc} % allow utf-8 input
\usepackage[T1]{fontenc}    % use 8-bit T1 fonts
\usepackage[bookmarks=false]{hyperref}       % hyperlinks
\usepackage{url}            % simple URL typesetting
\usepackage{booktabs}       % professional-quality tables
\usepackage{amsfonts}       % blackboard math symbols
\usepackage{nicefrac}       % compact symbols for 1/2, etc.
\usepackage{microtype}      % microtypography
\usepackage{xcolor}         % colors
\usepackage{caption}
\usepackage{amsmath}
\usepackage{amsthm}
\usepackage{subcaption}
\usepackage{enumitem}
\usepackage{bigstrut}
\usepackage{multicol}
\usepackage{multirow}
\usepackage{xspace}
\usepackage{amssymb}
\usepackage{dsfont}
\usepackage[ruled]{algorithm2e}

\definecolor{beaublue}{rgb}{0.84, 0.9, 0.95}
\definecolor{blackish}{rgb}{0.2, 0.2, 0.2}

\definecolor{beaublue2}{rgb}{0.84, 0.9, 0.95}
\definecolor{blackish2}{rgb}{0.2, 0.2, 0.2}

\makeatletter
\newcommand\fs@nobottomruled{\def\@fs@cfont{\bfseries}\let\@fs@capt\floatc@ruled
  \def\@fs@pre{}% \hrule height.8pt depth0pt \kern2pt
  \def\@fs@post{}% Formerly \def\@fs@post{\kern2pt\hrule\relax}%
  \def\@fs@mid{\kern2pt\hrule\kern2pt}%
  \let\@fs@iftopcapt\iftrue}
\makeatother

\newtheorem{theorem}{Theorem}

\makeatletter
\DeclareRobustCommand\onedot{\futurelet\@let@token\bmv@onedotaux}
\def\bmv@onedotaux{\ifx\@let@token.\else.\null\fi\xspace}
\def\eg{\emph{e.g}\onedot}

 \def\vs{\emph{vs}\onedot}
\def\wrt{w.r.t\onedot}

\def\bigoh{\mathcal{O}}
\makeatother

\title{REFINE: Random RangE FInder for Network Embedding}

\author{%
  Hao Zhu \\
  Australian National University\\
  Data 61/CSIRO\\
  \texttt{allenhaozhu@gmail.com} \\
  \And
  Piotr Koniusz\thanks{The corresponding author. The code is available at: {\color{red}{\url{https://github.com/allenhaozhu/REFINE}}}.\vspace{-0.5cm}} \\
Data 61/CSIRO\\
  Australian National University \\
  \texttt{name.surname@data61.csiro.au} \\
}

\begin{document}

\maketitle

\begin{abstract}% todo: not match the title
Network embedding approaches have recently attracted considerable interest as they  learn low-dimensional vector representations of nodes. Embeddings based on the matrix factorization are effective but they are usually computationally expensive due to the eigen-decomposition step. In this paper, we propose a Random RangE FInder based Network Embedding (REFINE) algorithm, which can perform  embedding on one million of nodes (YouTube) within 30 seconds in a single thread. REFINE is $10\times$ faster than ProNE, which is $10-400\times$ faster than other methods such as LINE, DeepWalk, Node2Vec, GraRep, and Hope. Firstly, we formulate our network embedding approach as a skip-gram model, but with an orthogonal constraint, and we reformulate it into the matrix factorization problem. Instead of using randomized tSVD (truncated SVD) as other methods, we employ the Randomized Blocked QR decomposition to obtain the node representation fast. Moreover, we design a simple but efficient spectral filter for network enhancement to obtain higher-order information for node representation. Experimental results prove that REFINE is very efficient on datasets of different sizes (from thousand to million of nodes/edges) for node classification, while enjoying a good performance.
\end{abstract}

\section{Introduction}
Network embeddings have drawn a lot of interest due to their ability to produce low-dimensional representation for nodes while encapsulating the structure/properties of the network~\cite{cui2018survey}. 
%
%Such representations  serve as latent features and  benefit  off-the-shelf machine learning  for a variety of tasks on graphs \eg, node classification~\cite{perozzi2014deepwalk}, link prediction~\cite{grover2016node2vec} and network reconstruction~\cite{wang2016structural}.
Such representations  serve as latent features for  off-the-shelf machine learning  for a variety of tasks on graphs \eg, node classification~\cite{perozzi2014deepwalk,pmlr-v115-sun20a,zhu2021graph}, link prediction~\cite{grover2016node2vec}, graph classification \cite{kon_tpami2020a}, relation learning \cite{Wang_2020}, trajectory analysis \cite{Prabowo_2019,shao2019flight,shao2021predicting}, spatio-temporal graphs \cite{kon_eccv16,kon_tpami2020b} and network reconstruction~\cite{wang2016structural}.

Many  network embedding approaches are based on random walks~\cite{perozzi2014deepwalk}, matrix factorization~\cite{qiu2018network} and deep learning~\cite{wang2016structural}. 
The skip-gram model has significantly advanced network embeddings \eg,  DeepWalk~\cite{perozzi2014deepwalk}, LINE~\cite{tang2015line}, PTE~\cite{tang2015pte}, and Node2Vec~\cite{grover2016node2vec} approaches, which  sample node pairs from $k$-step transition matrices with different values of $k$, and  train a skip-gram~\cite{mikolov2013efficient} model on these pairs to get node embeddings. 
The above methods can be unified into the closed-form Matrix Factorization (MF) framework \cite{qiu2018network} with two steps: (i) building a higher-order proximity matrix and (ii) obtaining the node embedding by using eigen-decomposition. 
Although MF provides an efficient  way to obtain network embeddings compared with skip-gram based methods, it faces the computational and space challenges on large scale networks ($\geq 10,000$ nodes) as these two steps  depend on SVD  and dense  proximity matrices.  

Many methods  use different approximation approaches to accelerate the MF based network embedding because the SVD, especially on dense matrices, limits the ability to scale up the efficient network embedding. 
Some methods  bypass higher-order proximity computations~\cite{zhang2019prone}, or  sparsify higher-order proximity matrices~\cite{qiu2019netsmf} \eg,  they use the randomized SVD~\cite{halko2011finding}  to achieve acceleration. 
Although \cite{zhang2019prone,qiu2019netsmf} are very fast  network embedding approaches based on matrix factorization, they depend on SVD, which limits their use on large scale networks.
Therefore, approach \cite{zhang2018billion} employs random projections, a simple and powerful technique, which forms a low-dimensional embedding space for the network while preserving the original graph structure. 
Although network embeddings based on random projections are very efficient, the performance is worse compared to learning-based approximation methods \cite{zhang2019prone,qiu2019netsmf}.

We present a new method to offer a trade-off between speed \cite{zhang2018billion} and performance  \cite{zhang2019prone,qiu2019netsmf}. By adding an orthogonal constraint on context vectors, we propose another Skip-Gram Network Embedding (SGNE) framework.  Subsequently, %we show that the corresponding  matrix factorization can be solved without SVD % to obtain embeddings. Instead of SVD, we use
%by a range finder to estimate the low-rank approximation. 
%Specifically, 
we obtain obtain node representations by a Randomized Blocked QR with power iteration (range finder), which is much faster than SVD-based methods. Finally, we present an efficient and effective spectral filter to enhance network embeddings by obtaining higher-order structural information of neighborhoods. Our contributions are:
\renewcommand{\labelenumi}{\roman{enumi}.}
%\vspace{-0.1cm}
\hspace{-1.0cm}
\begin{enumerate}[leftmargin=0.6cm]
    \item We propose a new skip-gram network embedding by adding  orthogonal constraints, making it an MF problem free of SVD.
    \item We propose a computationally more efficient than tSVD range finder based on the Randomized Blocked QR with power iteration.%to gain node representation.
    \item We propose a compact and effective spectral filter to enhance node representations.
    \item We validate the effectiveness/performance of REFINE on several standard benchmarks, and show it yields  network embeddings on one million of nodes within 30 seconds (a single CPU thread), which is  $\geq 10\times$ faster compared to the state of the art.
\end{enumerate}

%\vspace{-0.2cm}
\section{Related Work}
Many off-the-shelf ML techniques leverage network embeddings, which we review below. 
%We also discuss  Network Embedding Enhancement \cite{yang2017fast}.
%
%Deepwalk
 A popular deep embedding model called DeepWalk~\cite{perozzi2014deepwalk} uses truncated random walks (to explore the network structure) and the skip-gram word embedding model \cite{mikolov2013efficient} (to obtain embedding vectors of nodes).  LINE~\cite{tang2015line}  sets the walk length as one, and it introduces the negative sampling strategy~\cite{mikolov2013efficient} to accelerate training.  Node2Vec~\cite{grover2016node2vec} generalizes the above two methods and modifies the definition of neighborhood. The above  methods %based random walks 
 are  equivalent to factorizing a higher-order proximity matrix~\cite{qiu2018network}. 

%Matrix Factorization
Many explicit matrix factorization methods have been used for network embeddings.  GraRep~\cite{cao2015grarep} applies SVD to preserve higher-order proximity matrices (time complexity $\bigoh{(|V|^3)}$. HOPE~\cite{ou2016asymmetric} uses generalized SVD to preserve the asymmetric transitivity in directed networks. Community structure (a mesoscopic structure of network) is preserved by non-negative matrix factorization in~\cite{wang2017community}. 
Approach \cite{chen2017fast} uses the matrix factorization with sparsification  to accelerate SVD. Another approximate matrix factorization technique is used by approach \cite{yang2017fast}.  AROPE~\cite{zhang2018arbitrary} improves upon the above works by preserving arbitrary-order proximity simultaneously.

Many methods accelerate factorization \eg, ProNE \cite{zhang2019prone}  avoids computing the higher-order proximity matrix by initializing embeddings with a low-order proximity matrix and  applying a graph filter to improve the performance. 
Approach \cite{qiu2019netsmf} uses higher-order proximity  and then employs random-walk polynomial sparsification to higher-order proximity matrix. % and thus get an approximation. 
The above methods use a sparse proximity matrix with randomized tSVD which is much faster  in obtaining the network embedding than SVD on a dense matrix.

\begin{figure}
     \centering
     \begin{subfigure}[b]{0.48\textwidth}
         \centering
         \includegraphics[width=\textwidth]{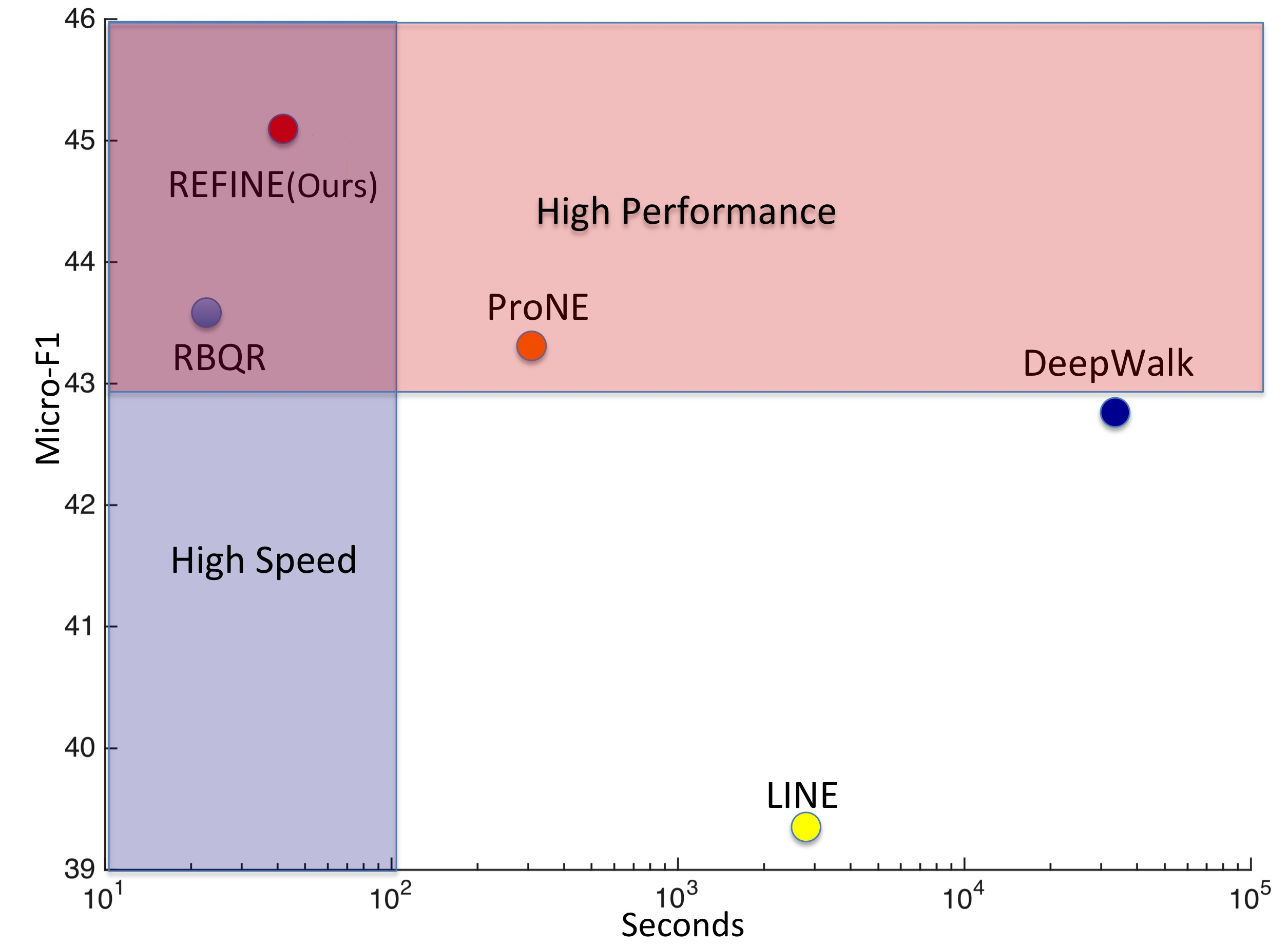}
         \caption{Micro-F1}
         \label{fig:micro}
     \end{subfigure}
     \begin{subfigure}[b]{0.48\textwidth}
         \centering
         \includegraphics[width=\textwidth]{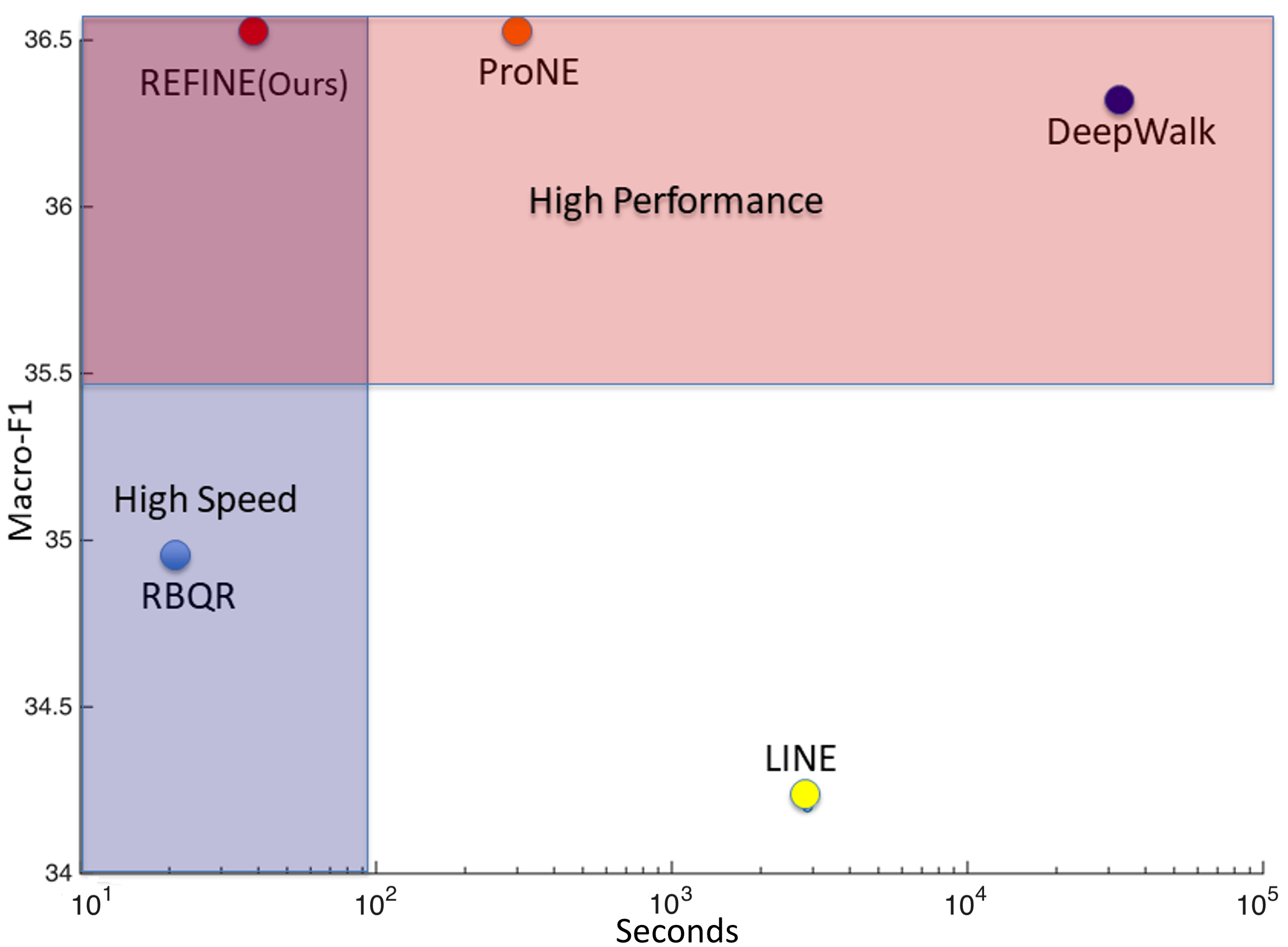}
         \caption{Macro-F1}
         \label{fig:macro}
     \end{subfigure}     
     \caption{The Micro-F1 (and Macro-F1) \vs runtime for different methods on the YouTube dataset (90\% of dataset used for training).}
\end{figure}
% \begin{figure}
% \includegraphics[width=0.48\textwidth]{micro.png}
% \includegraphics[width=0.48\textwidth]{macro.png}
% \caption{The Micro-F1 (and Macro-F1)\vs runtime for different methods on the YouTube dataset (90\% of dataset used for training).}
% \label{fig:micro}
% \end{figure}

\section{Methodology}
Below, we present our approach. Firstly, we propose a new objective function for SGNE, by imposing an orthogonal constraint on context vectors. We rewrite the objective function as the matrix factorization. Secondly, we present a range estimator based Randomized Blocked QR with power iterations to solve the matrix factorization.% for network embedding. 
Finally, to enhance network embedding with higher-order proximity, we propose a simple spectral filter, approximated  by the second-order Taylor expansion.

\subsection{SGNE as Column Pivoting QR Factorization }
Based on the Skip-Gram with Negative Sampling (SGNS) model\cite{mikolov2013efficient,zhang2019prone}, we introduce an orthogonal constraint on context vectors, which  leads to a very fast optimization procedure. Firstly, consider the  objective function:
\begin{equation}
\!\!\!\!\!l =-\!\!\!\!\sum_{(i,j)\in\hat{E}}[p_{i,j} \ln \sigma(\mathbf{r}_i^\top \mathbf{c}_j) + \lambda \phi({\hat{E},j}) \ln \sigma(-\mathbf{r}_i^\top \mathbf{c}_j)], \text{ s.t. } \mathbf{C}^\top\mathbf{C} = \mathbf{I},
\label{eq:SGNE2}
\end{equation}
where $\lambda\geq0$ is a coefficient controlling the negative noise sample ratio, $\sigma(\cdot)$ is the sigmoid function, context vector $\mathbf{c}_j$ is the $j$-th row of context matrix $\mathbf{C}\in \mathbb{R}^{n\times k}$, whereas $\mathbf{r}_i\in\mathbb{R}^k$ are $k$-dimensional embeddings and $p_{i,j}$ is the $(i,j)$-th coefficient of the degree-normalized adjacency matrix. Finally, $\phi({\hat{E},j})$ forms the empirical context for node $j$  %background noise of 
given the edge set  $\hat{E}$ associated with $j$, given as:
\begin{align}
    \phi({\hat{E},j})= \frac{\sum_{i:(i,j)\in\hat{E}} p_{i,j}}{\sum_{(i',j')\in\hat{E}} p_{i',j'}}.
\end{align}
Without the constraint $\mathbf{C}^\top\mathbf{C}=\mathbf{I}$,  Eq. ~\eqref{eq:SGNE2} has a trivial solution $c_{ij}\rightarrow\infty$. The constraint fulfills two different roles: i) it bounds context vectors to be normalized to the $\ell_2$ unit norm, and (ii) it decorrelates them. Thus, $\mathbf{C}$ can be thought of as a subspace as $k\ll n$. 
% A new constraint, $\mathbf{C}^\top\mathbf{C}=\mathbf{I}$,  makes the  optimization  efficient and prevents overfitting %to achieve a better generalization %to $\mathbf{c}_j$ because make%
% by making the context vectors decorrelated. 

A sufficient condition for minimizing the objective~\eqref{eq:SGNE2} is to let its partial derivative with respect to $\mathbf{c}^\top_j\mathbf{r}_i$ be equal zero, thus:
\begin{equation}
    \mathbf{r}_i^\top \mathbf{c}_j = \ln \frac{p_{i,j}}{\lambda \phi({\hat{E},j})},\quad (i,j)\in\hat{E}.
\end{equation}
If  $M_{ij} = \mathbf{r}_i^\top \mathbf{c}_j$,  $\mathbf{M}$ would result in a dense matrix, the source of inefficiency in MF.  Thus, ProNE \cite{zhang2019prone} defines a matrix $\mathbf{M}$ with entries:
%\vspace{-0.2cm}
\begin{equation}
M_{i,j}=\left\{
\begin{aligned}
&\ln \frac{p_{i,j}}{\lambda \phi({\hat{E},j})},&\quad (i,j)\in\hat{E} \\
&0,&\quad (i,j)\not\in\hat{E}. 
\end{aligned}
\right.
\end{equation}
Thus, %the objective function are transformed to 
we obtain an approximate  sparse matrix $\mathbf{M}$ with the low-rank product of  embedding matrix $\mathbf{R}$ and the context embedding matrix $\mathbf{C}$. %PRONE \cite{zhang2019prone} 
We avoid the truncated Singular Value Decomposition (tSVD), and achieve the optimal rank $k$ factorization \wrt the $p'$ norm by solving:
%\vspace{-0.3cm}
\begin{equation}
    \min_{\mathbf{R},\mathbf{C}} \|\mathbf{M}-\mathbf{M}^*\|_{p'}, \text{ s.t. } \mathbf{M}^* = \mathbf{R} \mathbf{C}^\top\!\!,\;\mathbf{C}^\top\mathbf{C} = \mathbf{I},
    \label{eq:REFINE}
\end{equation}
where $\mathbf{R}$ and $\mathbf{C}$ are $n\times k$ matrices whose  rows stand for a node embedding and context embedding, respectively. Note that this objective is different from other objectives in \cite{zhang2019prone,qiu2019netsmf,qiu2018network}. Intuitively,  through SVD, the matrix $\mathbf{M}$ can be decomposed into $\mathbf{M} = \mathbf{U}\boldsymbol{\Sigma}\mathbf{V}^\top$, where $\mathbf{U} = [\mathbf{u}_1,\cdots,\mathbf{u}_k]$ is orthogonal and $\boldsymbol{\Sigma} = \text{diag}([\sigma_1,\cdots,\sigma_k])$  with $\sigma_1 \ge \sigma_2 \ge \cdots\ge \sigma_k$. Thus, $\mathbf{R} = \mathbf{U}\boldsymbol{\Sigma}$ and $\mathbf{C} = \mathbf{V}$. Thus, we avoid SVD as its computational cost is  high.

\subsection{Randomized Range Finder for Network Embedding}
Another solution to Eq.~\eqref{eq:REFINE} can be obtained by the Randomized Range Finder, which  yields an orthonormal matrix $\mathbf{C}$ with very few columns, such that $\|(
\mathbf{I}-\mathbf{C}\mathbf{C}^\top)\mathbf{M})\|_F\le\epsilon$ for a desired tolerance $\epsilon$. Below, we explain the use of the randomized blocked QR factorization  to obtain $\mathbf{C}$. 

%\subsubsection{Randomized Blocked QR Factorization based Network Embedding}

Suppose that we seek a basis for the range of  matrix $\mathbf{M}$ with an exact rank $k$. Draw a random vector $\boldsymbol{\omega}$, and form the product $\mathbf{y} = \mathbf{M}\boldsymbol{\omega}$. 
%For now, the precise distribution of the random vector is unimportant; just think of $y$ as a random sample from the range of $M$. 
Repeat such a sampling process $k$ times: $\mathbf{y}_i = \mathbf{M}\mathbf{\omega}_i, i = 1, 2,\cdots, k$. Owing to the randomness, the set $\boldsymbol{\Omega} =[\boldsymbol{\omega}_1,\boldsymbol{\omega}_2,\cdots,\boldsymbol{\omega}_k]$ of random vectors is likely to be in the so-called general linear position. In particular, the random vectors form a linearly independent set and no linear combination falls in the null space of $\mathbf{M}$. As a result, the set $\mathbf{Y}=[\mathbf{y}_1,\cdots,\mathbf{y}_k]$ of sample vectors is also linearly independent, so it spans the range of $\mathbf{M}$. Therefore, to produce an orthonormal basis $\mathbf{C}$ for the range of $\mathbf{M}$, one just needs to orthonormalize the sample vectors (QR decomposition).
% The pseudocode for this algorithm is outlined in Algorithm~\ref{alg:RBI}. Of the parameters of the algorithm, $k$ (target rank) is problem dependent, while $q$ (power iteration numbers) is chosen by the user to control the quality and computational cost of the approximation. 
% The theoretical performance guarantees of randomized QR decomposition based low-rank approximation are given in the Theorem~\ref{theorem:lowrank}. 
The theoretical performance guarantees of randomized QR decomposition based low-rank approximation are given in the Theorem~\ref{theorem:lowrank}. 

The randomized QR decomposition works well for matrices whose singular values exhibit some decay, but it may produce a poor basis if the input matrix has a flat spectrum or the input matrix is very large. 
The power iteration is thus considered in our solution despite requiring extra $q$ times matrix–matrix multiplications ($\mathbf{M}^q\boldsymbol{\Omega}$) as the power iteration is far more accurate if  singular values of $\mathbf{A}$ decay slowly. 
A  heuristic we use has a nice property: if the original scheme ($q=1$) produces a basis whose approximation error is within a factor $c'$ of the optimum~\cite{halko2011finding}, the power scheme yields an approximation error within $c'^{1/q}$ of the optimum. The power iteration shrinks the approximation gap towards one exponentially fast. 

\begin{theorem}Based on \cite{halko2010finding}, 
let $\mathbf{M}$ be a  matrix ($m \times n$) of real numbers with singular values $\sigma_1 \ge \sigma_2 \ge \sigma_3 \ge \cdots$. Choose a target rank $k \ge 2$ and
an oversampling parameter $p \ge 2$, where $k + p = l$ and $l \le \min\{m, n\}$. Draw coefficients of matrix $\boldsymbol{\Omega}$ of size $n \times l$ according to the Normal distribution, $\Omega_{ij}\sim\mathcal{N}(0, 1/k)$, and construct the sample matrix $\mathbf{Y}(\boldsymbol{\Omega}) = \mathbf{M}\boldsymbol{\Omega}$. Then the expected approximation  error of $\text{\em QR}(\mathbf{Y}(\boldsymbol{\Omega}))\!=\!\mathbf{C}({\boldsymbol{\Omega}})$ yields:
\begin{equation}
\mathbb{E}_{\boldsymbol{\Omega}%: \Omega_{ij}\sim\mathcal{N}(0, 1/k)
}\left\|\mathbf{M}-\mathbf{C}({\boldsymbol{\Omega}})\mathbf{C}({\boldsymbol{\Omega}})^\top\mathbf{M}\right\|_{F} \leq \min\limits_{k+p=l}\bigg(1+\frac{k}{p-1}\bigg)^{\frac{1}{2}}\!\bigg(\sum_{j>k} \sigma_{j}^{2}\bigg)^{\frac{1}{2}}.
\end{equation}
\label{theorem:lowrank}
\end{theorem}

\vspace{-0.3cm}
Although the eigen-decomposition or randomized SVD can be avoided, the orthogonalization (QR decomposition) is the most computationally intensive part of the entire algorithm. 
Take QR decomposition as an example, the computational cost is approximately $2nk^2-\frac{2}{3}k^3$ flops. 
The cost is highly dependent on $k$. By blocking techniques, one can improve computational efficiency to obtain cost $2nb^2-\frac{2}{3}qb^3$ flops, where $b\ll k$.
%
% However, at the same time, the power iteration cannot efficiently put in the blocked QR \cite{martinsson2016randomized}, which will introduce another two QR decomposition for each iteration. 
% %
% In this paper, for the trade off between efficiency and effectiveness, we propose a network embedding algorithm based on Randomized Blocked Krylov QR. 
%
% Instead of using power iteration ($\mathbf{M}^{q-1}\mathbf{\Omega}$) for $\mathbf{M}$, krylov subspace builds the
% Krylov subspace in $q$ iterations as $K_q (A, V, q) = span\{\mathbf{\Omega}, \mathbf{M}\mathbf{\Omega},... , \mathbf{M}^{q-1}\mathbf{\Omega\}}$.
%

In the proposed by us randomized blocked QR based network embedding, we use power iterations to improve the performance and speed. In the experimental section, we  discuss the parameters for our techniques.
The pseudocode for our algorithm is outlined in Algorithm~\ref{alg:RBQR}. 
Of the parameters of the algorithm, $k$ (target rank) is problem dependent, whereas $b$ (block size) and $q$ (the number of power iterations) are chosen by the user to control the quality and computational cost of the approximation. The algorithm requires the choice of $b$ and $q$ to satisfy $qb \ge k$, 
\begin{algorithm}[b]
\SetAlgoLined
%\KwResult{Write here the result }
\KwIn{$\mathbf{M}\in \mathbb{R}^{n\times n}, \text{rank}~k\le n,b, q$}
\KwOut{$\mathbf{R}\in \mathbb{R}^{n\times k}$}
% $\mathbf{\Omega}\sim\mathbb{N}(0,1)^{n\times b}$\;
% $\mathbf{R}=[]$\;
% $\mathbf{C}=[]$\;
 \For{$i \gets 1$ \textbf{to} $k/b$} {
%   \If{$a_i > max$} {
%     $max \gets a_i$\;
%   }
    $\boldsymbol{\Omega}:\Omega_{ij}\sim\mathcal{N}(0,1)\quad$ ($\boldsymbol{\Omega}$ is of ${n\times b}$ size)\\
    %$\mathbf{R} = \mathbf{A}^q\mathbf{\Omega}$\;
    % $\mathbf{Y} = -\mathbf{C}\mathbf{R}^\top \mathbf{\Omega}$ \;
    % $\Omega = \mathbf{M}\Omega$ \;
    % $\mathbf{Y} = \mathbf{Y} + \Omega$ \;
    $\mathbf{C}_i = \text{QR}(\mathbf{M}^q\boldsymbol{\Omega})$\\
    $\mathbf{C}_i = \text{QR}(\mathbf{C}_i-\sum_{j=1}^{i-1}\mathbf{C}_j \mathbf{C}_j^\top \mathbf{C}_i)$\\
    $\mathbf{R}_i^\top = \mathbf{C}_i^\top \mathbf{M}$\\
    %$\mathbf{M} = \mathbf{M}- C_i R_i$ \;
    % $\mathbf{R} = [\mathbf{R},R_i]$ \;
    % $\mathbf{C} = [\mathbf{C},C_i]$ \;
    
}
\Return{$\mathbf{R}=[\mathbf{R}_1,\cdots,\mathbf{R}_{k/b}]$}
% $\mathbf{K} = diag[\mathbf{A}\mathbf{R},\mathbf{A}^2\mathbf{R},...,\mathbf{A}^q\mathbf{R}]$\;
%  $\mathbf{C} = \textbf{qr}(\mathbf{K});\quad \mathbf{C}\in \mathbb{R}^{n\times bq}$\;
%  $\mathbf{Z} = \mathbf{A}\mathbf{C}$\;
\caption{Network Embedding based on the Randomized Blocked QR (RBQR).}
\label{alg:RBQR}
\end{algorithm}
and it is not the standard blocked QR as we avoid computing the residual  $\mathbf{M}'=\mathbf{M}-\mathbf{R}\mathbf{C}^\top\!$, which is a dense matrix leading to extra storage costs. Empirically, we found that removing this term does not influence the final performance. % by much. 

\subsection{Spectral Graph Filter for Network Embedding Enhancement}
The randomized blocked QR factorization %yields node representations, it 
relies on the low-proximity matrix, which means that such a node embedding does not capture the relationship between distant neighbors. 
In social networks, this is of concern as not direct neighbors are of  interest \eg, representing close friends and distant acquaintances.

To improve our model, we  capture such  relations via the graph diffusion.
Intuitively, we  apply the heat to the node under consideration and then continuously diffuse the heat towards other neighbors. After a certain time,  the heat distribution defines the edge weights from the starting node to other nodes.
Thus, we obtain a matrix that defines a new, continuously weighted graph. We define a graph diffusion on node embedding as:
\begin{equation}
    \mathbf{R}^*=\sum_{k'=0}^{K} \theta_{k'} \mathbf{T}^{k'}\mathbf{R},
\label{eq:polySpec}
\end{equation}
where $\theta_{k'}$ are coefficients and $\mathbf{T}$ is the transition matrix  $\mathbf{D}^{-1}\mathbf{A}$. Coefficients $\theta_{k'}$ are predefined by the specific diffusion variant we choose, such as the heat kernel~\cite{klicpera2019diffusion} or Markov diffusion kernel~\cite{zhu2021simple}. Increasing $K$ will utilize the information from the  $K$-hop neighborhoods of the node at an increased computational cost. We set $K=2$ for all datasets, as empirically we observe it is sufficient. 
%
%We will show this method is also efficient and significantly improve the network embedding in the experiment section.

% \begin{figure}
% \includegraphics[width=0.45\textwidth]{macro.png}
% \caption{The Macro-F1 vs runtime for different methods on the YouTube with 90\% training ratio.}
% \label{fig:macro}
% \end{figure}

\subsection{Computational Complexity}
We compare the computational cost of the randomized QR and Algorithm \ref{alg:RBQR}. To this end, let $C_{\text{spmm}}$,$C_{\text{mm}}$ and $C_{\text{qr}}$ denote the scaling constants for the cost of sparse matrix-matrix multiplication, matrix-matrix multiplication and a full QR factorization, respectively.  The computational complexity for the randomized QR is  $(q+1)C_\text{spmm}|E|k + C_\text{qr} nk^2$, where $|E|$ is the number of edges.
%Algo.\ref{alg:RBKI} is easily seen to be $2C_{mm}n^2k + C_{qr} nk^2$.
Alg.~\ref{alg:RBQR} costs $(q+1)C_\text{spmm}|E|k + C_\text{mm}nk^2 + \frac{2}{k/b}C_\text{qr}nk^2$, where $b$ is the block size. The blocked QR is faster if $n/b\ge2$, whereas more  power iterations (larger $q$) yields better results and costs more. Alg.~\ref{alg:RBQR} spends less time executing the full QR factorization, as expected. The computational cost of Eq.~\eqref{eq:polySpec} is $3|E|k + 3nk$. % like we mentioned the approximated implementation.

\begin{table*}[htbp]
  \centering
  \caption{Micro/Macro-F1(\%) of node classification on the Blogcatalog and the PPI datasets.}
  \resizebox{1\textwidth}{!}{
    \begin{tabular}{c|c|ccccc|ccccc|}
    \hline
    \multicolumn{1}{c|}{Metric} & ALG & \multicolumn{5}{c|}{BlogCatalog}      & \multicolumn{5}{c|}{PPI} \bigstrut\\
\cline{3-12}          &       & 10.0\% & 30.0\% & 50.0\% & 70.0\% & 90.0\% & 10.0\% & 30.0\% & 50.0\% & 70.0\% & 90.0\% \bigstrut\\
    \hline
    \multicolumn{1}{c|}{Micro-F1} & LINE  & 25.35 & 32.05 & 35.16 & 36.61 & 37.35 & 11.7  & 14.2  & 16    & 17.82 & 19.59 \bigstrut\\
          & DeepWalk & 35.85 & 39.91 & \textbf{41.62} & \textbf{42.45} & \textbf{42.9}  & 16.06 & 19.37 & 21.26 & 22.63 & 24.36 \\
          %& NetMF &  \textbf{38.33} &  \textbf{41.43} &  \textbf{42.67} &  \textbf{43.34} &  \textbf{43.15} &  \textbf{18.05} & {21.80} & {23.10} & {24.40} & {25.96} \bigstrut\\
\cline{2-12}     
          & ProNE & \textbf{36.52} & \textbf{39.97} & {41.20} & {41.75} & {42.16} & {17.37} & {22.45} & {24.28} & {25.08} & {26.31} \bigstrut\\
          & RBQR & {32.88} & {36.42} & {37.85} & {38.61} & {39.33} & {16.44} & {20.90} & {22.59} & {23.29} & {23.71} \bigstrut\\
          & REFINE & {36.46} & {39.75} & {41.00} & {41.75} & {42.29} & {17.79} &  \textbf{22.57} &  \textbf{24.30} &  \textbf{24.96} &  \textbf{26.35} \bigstrut\\
    \hline
    \multicolumn{1}{r|}{Macro-F1} & LINE  & 14.38 & 19.11 & 21.36 & 22.25 & 22.62 & 9.5   & 12.15 & 13.82 & 15.35 & 15.92 \bigstrut\\
          & DeepWalk & \textbf{21.16} & \textbf{25.59} & \textbf{27.58} & \textbf{28.47} &  \textbf{28.66} & 12.89 & 16.71 & 18.25 & 19.48 & 20.36 \\
          %& NetMF &  \textbf{23.12} &  \textbf{26.7} &  \textbf{28.31} &  \textbf{28.91} &  {28.44} &  \textbf{13.97} &  \textbf{18.29} &  \textbf{19.94} &  \textbf{21.01} &  \textbf{20.99} \bigstrut\\
\cline{2-12}          
          & ProNE & {18.04} & {22.68} & {24.05} & {24.95} & {25.03} & {12.42} & {17.44} & {19.59} & {20.52} & {20.86} \bigstrut\\ 
          & RBQR  & {13.74} & {18.25} & {20.21} & {21.30} & {21.56} & {11.10} & {15.69} & {17.97} & {18.89} & {18.71} \bigstrut\\   
          & REFINE & {17.76} & {22.61} & {24.17} & {25.09} & {25.35} & {12.67} & {17.57} & {19.73} & {20.60} & {20.86} \bigstrut\\
    \hline
    \end{tabular}%
    }
  \label{tab:small}%
\end{table*}%

\begin{table*}[htbp]
  \centering
  \caption{Micro/Macro-F1(\%) of node classification on the Wikipedia and Flickr datasets.}
  \resizebox{1\textwidth}{!}{
    \begin{tabular}{c|c|ccccc|ccccc|}
    \hline
    \multicolumn{1}{c|}{{Metric}} & {ALG} & \multicolumn{5}{c|}{Wikipedia} & \multicolumn{5}{c|}{Flickr} \bigstrut\\
\cline{3-12} 
& \multicolumn{1}{c|}{} & 10.0\% & 30.0\% & 50.0\% & 70.0\% & 90.0\% & 1.0\% & 3.0\% & 5.0\% & 7.0\% & 9.0\% \bigstrut\\
    \hline
    \multicolumn{1}{c|}{{Micro-F1}} & LINE  & 41.3  & 48.35 & 51.89 & 53.57 & 54.86 & 25.3  & 28.64 & 30.07 & 31.28 & 32.34 \bigstrut\\
          & DeepWalk & 42.32 & 47.02 & 48.65 & 49.8  & 50.35 &  \textbf{32.06} &  \textbf{35.89} &  \textbf{37.46} &  \textbf{38.29} &  \textbf{38.84} \\
          %& NetMF & \textbf{50.1} &  \textbf{55.81} &  \textbf{57.26} &  \textbf{58.47} &  \textbf{59.13} & {31.97} & {35.07} & {36.24} & {36.82} & {37.19} \bigstrut\\
\cline{2-12}     
          & ProNE & {48.61}  & {53.96} & {55.63} & {56.52} & {57.43} & {30.77} & {35.14} & {36.69} & {37.54} & {38.12} \\
          & RBQR & {45.72}  & {49.73} & {50.88} & {51.56} & {52.31} & {30.55} & {34.23} & {35.58} & {36.41} & {37.04} \\
          & REFINE & \textbf{51.17}  & \textbf{56.15} & \textbf{57.50} & \textbf{58.34} & \textbf{58.84} & {31.17} & {35.22} & {36.74} & {37.66} & {38.32} \\
          \hline
    \multicolumn{1}{c|}{{Macro-F1}} & LINE  & 8.51  & 10.53 & \textbf{12.63} & \textbf{13.4}  & \textbf{13.16} & 9.01  & 13.42 & 15.77 & 17.44 & 18.68 \\
          & DeepWalk & 7.26  & 9.02  & 9.71  & 10.03 & 9.92  &  \textbf{13.36} &  \textbf{19.45} &  \textbf{22.21} &  \textbf{23.94} &  \textbf{25.07} \\
          %& NetMF & {9.04} &  \textbf{12.44} &  \textbf{13.67} &  \textbf{14.04} &  \textbf{14.91} & {12.32} & {17.32} & {19.42} & {20.68} & {21.57} \bigstrut\\
\cline{2-12}         
          & ProNE & 8.40  & 10.73  & 11.40 & 11.83 & 12.21 & 8.21 & 13.55 & 16.14 & 17.86 & 19.12 \\  
          & RBQR & 7.10  & 9.39  & 9.98 & 10.36 & 10.63 & 7.68 & 12.21 & 14.47 & 15.98 & 17.21 \\   
          & REFINE & \textbf{9.42}  & \textbf{11.75}  & 12.36 & 12.79 & 13.11 & 8.18 & 13.50 & 16.29 & 18.21 & 19.54 \\          
    \hline
    \end{tabular}%
    }
  \label{tab:large}%
\end{table*}%

\section{Experiments}
Below, we evaluate our method on the node classification. We study the computational efficiency, the performance and  parameters.

\begin{table}[b]
\centering
\caption{The statistics of datasets.}
%\resizebox{.48\textwidth}{!}{
\begin{tabular}{l|l|l|l|l|l|}
\hline
Dataset  & BlogCatalog & Wiki   & PPI   & Flickr   & YouTube \\
\hline
\#nodes  & 10312       & 4777   & 3890  & 80513  & 1138499 \\
\#edges  & 333983      & 184812 & 76584 & 5899882 & 2990443 \\
\#labels & 39          & 40     & 50    & 195     & 47     \\
\hline
\end{tabular}
%}
\label{tab:datasets}
\end{table}
\subsection{Experiments Setting}
%We evaluate the algorithms on five widely-used real-world network datasets: BlogCatalog~\cite{zafarani2009social}, Protein-Protein Interactions (PPI~\cite{grover2016node2vec}, Wikipedia, Flickr~ \cite{tang2009relational}, YouTube~\cite{zafarani2009social}.
We evaluate the algorithms on four widely-used real-world network datasets listed below:
\renewcommand{\labelenumi}{\arabic{enumi})}
% \vspace{-0.25cm}
\hspace{-1.0cm}
\begin{enumerate}[leftmargin=0.6cm]
    \item  \textbf{BlogCatalog}~\cite{zafarani2009social} is a network of social relationships of bloggers in the BlogCatalog website, whose labels represents interests of bloggers.
    \item Protein-Protein Interactions (\textbf{PPI})~\cite{grover2016node2vec} is a subgraph of the PPI network for Homo Sapiens, whose labels represent biological states. 
    \item  \textbf{Wikipedia} is a co-occurrence network of words appearing in the first million bytes of the Wikipedia dump. The labels are Part-of-Speech (POS) tags inferred using the Stanford POS-Tagger.
    \item \textbf{Flickr}~ \cite{tang2009relational} is a network of contacts between Flickr users, whose labels represent the user’s interest group. 
    \item \textbf{YouTube} \cite{zafarani2009social} is a social network between YouTube users. The labels represent groups of viewers that enjoy common video genres. The statistics of these datasets are shown in Table~\ref{tab:datasets}.
\end{enumerate}
Experiments were conducted on a Ubuntu workstation (AMD Ryzen 2700, 64G RAM).

\vspace{0.05cm}
\noindent\textbf{Baseline Algorithms.} 
We compare the proposed algorithms with LINE(2nd)~\cite{tang2015line} (number of samples 10, number of negative samples 5, initial learning rate 0.025), DeepWalk~\cite{perozzi2014deepwalk} (window size 10, walk length 40, and the number of walks 80), and ProNE~\cite{zhang2019prone}. 
The proposed methods include two variants, with or without spectral filters, referred to as RBQR (no spectral filter) and REFINE (RBQR with spectral filters). The dimension of representation  is set $k=128$.

For the node classification, we  follow the same experimental setting as DeepWalk. In particular, we randomly sample a portion of labeled nodes for training and use the rest for testing. For BlogCatalog, PPI and Wiki datasets, the training set size ranges from 10\% to 90\% of a given dataset, with 20\% step. For Flickr and YouTube, the training set size ranges from 1\% to 9\% of the dataset, with  2\% step. Using the one-vs-rest logistic regression, we repeat experiment  10 times and report the Micro-F1 and Macro-F1 performance.

\begin{table}[]
\centering
\caption{Micro-F1 scores for node classification. Asterisk `*' indicates that we use 9\% of a given dataset for training, other methods use 80\% of a given dataset for training.}
%\resizebox{.48\textwidth}{!}{
\begin{tabular}{l|l|l|l|l|l|l}
\hline
            & HARP  & RandNE & MILE  & NetSMF & LouvainNE & Ours  \\
\hline
Blogcatalog & 0.316 & 0.308  & 0.264 & 0.334  & 0.306     & \textbf{0.420}  \\
Flickr      & 0.384 & 0.385  & 0.386 & 0.356  & 0.389     & \textbf{0.389}* \\
YouTube     & 0.305 & 0.303  & 0.304 & 0.307  & 0.307     & \textbf{0.451}*\\
\hline
\end{tabular}
\label{tab:fast}
\end{table}

\subsection{Computational Efficiency}
%We compare the efficiency of different methods. 
%setting
The efficiency of baselines is accelerated by using 16 CPU threads, whereas our approach and ProNE  use a single CPU thread. Note that REFINE can be easily extended to a multi-threaded REFINE as QR scales gracefully for multi-threading compared with SVD.
As shown in Figure~\ref{fig:micro}, DeepWalk needs more than 10K seconds to obtain the node representation on  YouTube. LINE is $10\times$ faster than DeepWalk while ProNE is $10\times$ faster than LINE. 
Our approach with/without spectral filters is $10\times$ faster than ProNE. 
Table~\ref{tab:timecost} reports the running times (both I/O and computation time) of our approach and the fastest baseline, ProNE (other methods are much slower). %As other methods are much slower, we only compare our method  with  ProNE on other datasets. 
%
% Some matrix factorization baselines are much slower than them. For example, the time complexity of GraRep is $O(|V|^3)$, making it infeasible for relatively big networks, such as Youtube of 1.1 million nodes.
% %
% NetMF~\cite{qiu2018network} need to compute higher-order proximity matrix and make the graph very dense, making it also infeasible for large networks.
%
The runtime results suggest that for PPI and Wiki (small networks of 1,000+ nodes), REFINE  requires less than 0.15  seconds to complete while the fastest baseline ProNE is at least $10\times$ slower. 
Similar speed-ups are consistently observed on BlogCatalog and Flickr,  moderate-size networks (10K+ nodes), and YouTube, a relatively big network (1M+ nodes). 
Remarkably, our proposed method  embeds the YouTube network within 40 seconds by using one thread, whereas ProNE takes 300 seconds. Remaining baselines need between half an hour and a dozen of hours. Without spectral filters,  using RBQR alone (Algorithm~\ref{alg:RBQR} alone) to obtain the network embedding takes \textbf{26} seconds. 
%while by using 20 threads LINE costs 100 minutes, DeepWalk requires 19 hours, and node2vec takes more than fives days.
%
We analyzed the time cost of our method and found that the random generation costs $\geq$15\% of the total time. 
In practice, the cost may be limited by using pools of random numbers. In this paper, we kept the cost of random generator in our comparisons. 

\begin{tcolorbox}[width=1.0\linewidth, colframe=blackish, colback=beaublue, boxsep=0mm, arc=3mm, left=1mm, right=1mm, right=1mm, top=1mm, bottom=1mm]
To summarize, the single-threaded REFINE is about 8-20$\times$ faster than the ProNE, which is 10–400x faster than the 16-threaded LINE, DeepWalk, and Node2Vec.
\end{tcolorbox}

\begin{table}[]
\centering
\caption{Computational efficiency (runtime in seconds).}
% \resizebox{.48\textwidth}{!}{
\begin{tabular}{l|l|l|l|l|l}
\hline
Dataset  & BlogCatalog & Wiki   & PPI   & Flickr   & YouTube \\
\hline
ProNE & 10.04          & 2.06     & 0.959    & 195     & 305 \\
REFINE  & \textbf{0.49}            & \textbf{0.15}     & \textbf{0.097}    & \textbf{13}    & \textbf{40} \\
\hline
\end{tabular}
%}
\label{tab:timecost}
\end{table}

\subsection{Performance}
For YouTube, Figure~\ref{fig:micro} shows that  REFINE enjoys  the best Micro-F1 score and the best runtime compared to three fastest baselines, given training set of 90\% size of the dataset.  
 Figure~ \ref{fig:macro} shows that REFINE simultaneously achieves the best Macro-F1 score and the best runtime compared with competing methods. 
Thus, REFINE without spectral filtering  outperforms other more advanced methods such as DeepWalk on YouTube.
%Table \ref{tab:you} provides further results on  YouTube.
%
%Due to the space limit, we do not  provide Macro-F1 results on YouTube but the trend is similar as for  Micro-F1 scores.
%
 Tables \ref{tab:small} and \ref{tab:large} summarize the prediction performance on small- and large-scale datasets, respectively, and report the results in terms of Micro-F1 and Macro-F1 metrics.  
We provide the embedding results generated by the RBQB (Algorithm~\ref{alg:RBQR}) without spectral filters as well as results with the spectral filter (REFINE). 
We note that REFINE, RBQR, LINE, and ProNE have all similar performance in Macro-F1 because they are based on the low-proximity matrix factorization. % (less than second-order).
Benefiting from the high-order information, DeepWalk performs well in terms of the Macro-F1 score. 
Table  \ref{tab:small} shows that REFINE works very well in terms of the Micro-F1 scores outperforming DeepWalk by $\sim$8\% on Wikipedia, and outperforming LINE for  training ratios of 10\% and 30\% on Macro-F1. %
REFINE also outperforms the initial RBQR (Algorithm~\ref{alg:RBQR}) by a large margin.  We also conduct the comparison with fast network embedding methods without matrix factorization~\cite{bhowmick2020louvainne,liang2018mile,chen2018harp}. As shown in Table~\ref{tab:fast}, our method outperforms other methods significantly. Kindly note that on Flickr and Youtube, our method only uses 9\% of a given dataset for the training set, whereas other methods use 80\% of a given dataset for training. 

% We observe that ProNE consistently generates better results than baselines across five datasets, demonstrating its strong effectiveness.
% %
% Interestingly, it turns out that the simple sparse matrix factorization (SMF) step for fast embedding initialization in ProNE is comparable to or sometimes even better than existing popular network embedding benchmarks.
% %
% With the spectral propagation technique further incorporated, ProNE generates the best performance among all baselines due to its effective modeling of local structure smoothing and global clustering information.

\subsection{Parameter Analysis}
Our approach has two parameters which control the efficiency and effectiveness, that is, power iteration number $q$ and the block size $b$. As we aforementioned, the power iteration can improve the accuracy of the low-rank approximation at the cost of extra computations. The block size, given a rank $k$, can determine the number of blocks. Below, we analyze the computational cost of blocked QR, which is highly dependent on the block number.

\begin{figure}[!htbp]
\centering
\includegraphics[width=0.8\textwidth]{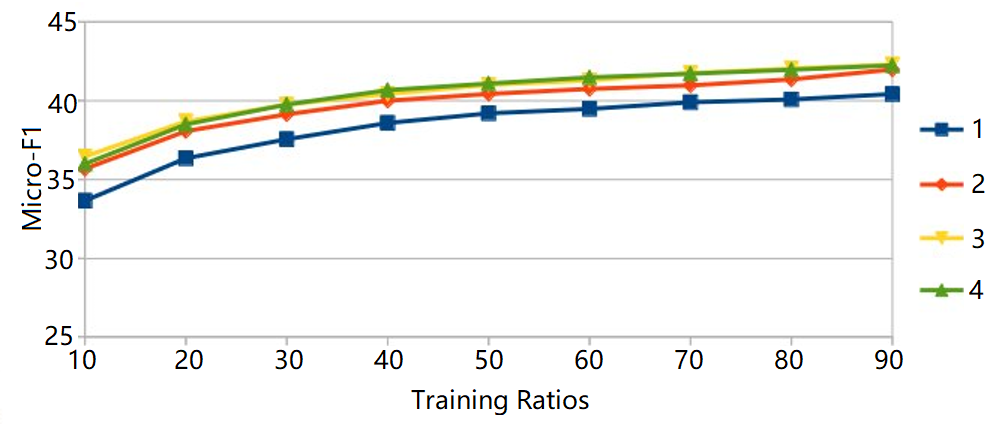}
\caption{Micro-F1 \wrt the number of  power iterations on  Blogcatalog under different training ratios in range 10\%-90\%.}
\label{fig:PI}
\end{figure}

\vspace{0.1cm}
\noindent\textbf{Power Iteration.} Figure~\ref{fig:PI}  shows Micro-F1 \wrt the number of  power iterations (ranging from 1 to 4, indicated by the four curves) on Blogcatalog. The plot shows that the larger number of iterations is, the better the results of REFINE are.  REFINE with $q\geq2$ significantly improves the performance over without power iteration ($q=1$). The performance at  $q=3$  saturates, whereas  $q=4$ does not offer any further significant gain.

\begin{figure}
\centering
\includegraphics[width=0.8\textwidth]{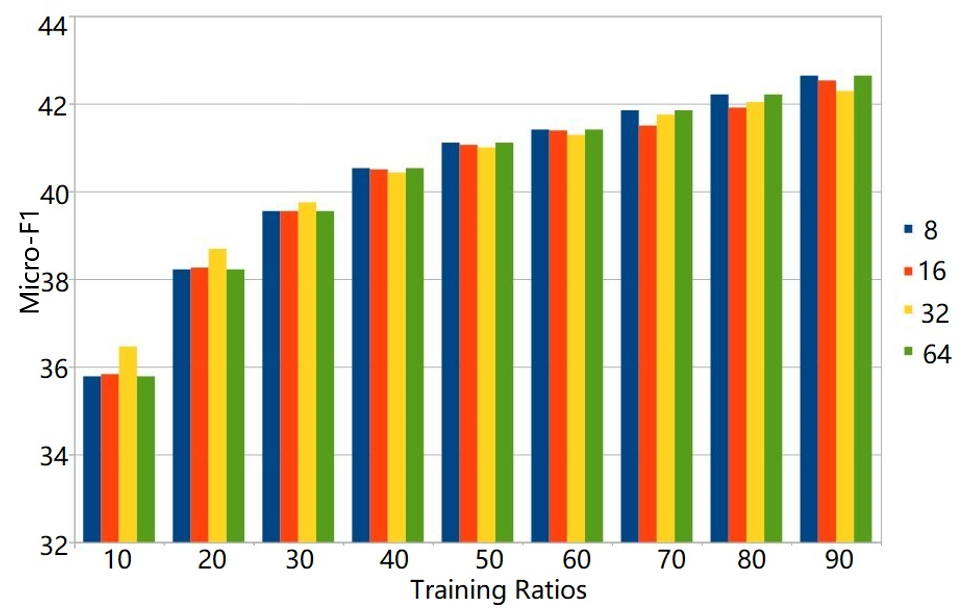}
\caption{Micro-F1 \wrt different block sizes on  Blogcatalog under different training ratios (10-90\%).}
\label{fig:block}
\end{figure}

\vspace{0.1cm}
\noindent\textbf{Block Size.} Figure~\ref{fig:block} shows Macro-F1 \wrt different block sizes (8,16, 32, 64, indicated by different colors) on the Blogcatalog dataset. The plot shows that there are no significant differences between these variants. For the sake of efficiency, the smaller the block size is, the lesser the computational cost is \eg, the QR decomposition may account for less than 10\% of runtime of the whole algorithm, which appears to be faster than the cost of random generation. 

\section{Conclusions}
In this work, we have proposed REFINE, a fast and scalable network embedding approach. 
REFINE achieves a favourable computational efficiency and performance compared to recent powerful network embedding approaches, such as DeepWalk and LINE.
%, node2vec, and HOPE.
%
The proposed method is $8-20\times$ faster than the ProNE, which is $10-400\times$ faster than the aforementioned baselines that are already accelerated by multi-threaded codes. 
Due to the speed and accuracy of REFINE, its use is suitable for scenarios where node embeddings are frequently updated \eg, in commercial recommendation systems.

\section*{Acknowledgement}
This research is supported by the Australian Government Research Training Program (RTP) scholarship.
% In this paper, we mainly introduce the detail of our method in the classification problem.
% %
% This method is easily extended to other applications, such as regression problems.
% %
% In addition, this method is also very easy to integrate the features provided by the server to model collaboratively because the server-side parameters are stored in plaintext. 
% %
% The approach mentioned in this article is only to clarify the problem in a simplified model and to presend a solution in a few applications. We will extend it in future work.
\bibliography{neurips_2020}
\bibliographystyle{plain}

\end{document}